\documentclass[sigconf]{acmart}
\usepackage{cleveref}

\usepackage{booktabs}
\usepackage{multirow}
\usepackage{xcolor}

\usepackage{pifont}  
\usepackage{graphicx}
\usepackage{tabularx}
\usepackage{adjustbox}
\usepackage{hyperref}
\newcommand{\cmark}{\textcolor{green!60!black}{\checkmark}}  
\newcommand{\xmark}{\textcolor{red}{\ding{55}}}  

\newcolumntype{L}[1]{>{\raggedright\arraybackslash}p{#1}}  
\newcolumntype{Y}{>{\raggedright\arraybackslash}X} 

\renewcommand\footnotetextcopyrightpermission[1]{} 
\AtBeginDocument{%
  }

\begin{document}

\title{
    \raisebox{-.5ex}{\llap{\includegraphics[width=0.7cm]{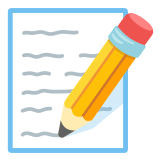}}}\,
    \textsc{NoTeS-Bank}: Benchmarking Neural Transcription and Search for Scientific Notes Understanding
}



\author{Aniket Pal}
\authornote{These authors are Joint First Authors and contributed equally.}
\affiliation{%
  \institution{CVIT Lab, IIIT Hyderabad}
  \city{Hyderabad}
  \state{Telangana}
  \country{India}
}

\author{Sanket Biswas}
\authornotemark[1]
\affiliation{%
  \institution{Computer Vision Center, UAB}
  \city{Barcelona}
  \country{Spain}
}

\author{Alloy Das}
\authornotemark[1]
\affiliation{%
  \institution{Habitat Labs, Habitat Lens Pvt. Ltd.}
  \city{Kolkata}
  \state{W. B.}
  \country{India}
}

\author{Ayush Lodh}
\authornote{These authors are Joint Second Authors and provided pivotal support.}
\affiliation{%
  \institution{Habitat Labs, Habitat Lens Pvt. Ltd.}
  \city{Kolkata}
  \state{W. B.}
  \country{India}
}

\author{Priyanka Banerjee}
\authornotemark[2]
\affiliation{%
  \institution{Habitat Labs, Habitat Lens Pvt. Ltd.}
  \city{Kolkata}
  \state{W. B.}
  \country{India}
}

\author{Soumitri Chattopadhyay}
\authornotemark[2]
\affiliation{%
  \institution{UNC Chapel Hill}
  \city{Chapel Hill}
  \state{NC}
  \country{USA}
}

\author{Dimosthenis Karatzas}
\authornote{These authors are the Directors of the whole Project}
\affiliation{%
   \institution{Computer Vision Center, UAB}
  \city{Barcelona}
  \country{Spain}
}

\author{Josep Lladós}
\authornotemark[3]
\affiliation{%
   \institution{Computer Vision Center, UAB}
  \city{Barcelona}
  \country{Spain}
}

\author{C.V. Jawahar}
\authornotemark[3]
\affiliation{%
  \institution{CVIT Lab, IIIT Hyderabad}
  \city{Hyderabad}
  \state{Telangana}
  \country{India}
}

\renewcommand{\shortauthors}{Pal et. al.}

\begin{abstract}
  Understanding and reasoning over academic handwritten notes remains a challenge in document AI, particularly for mathematical equations, diagrams, and scientific notations. Existing visual question answering (VQA) benchmarks focus on printed or structured handwritten text, limiting generalization to real-world note-taking. To address this, we introduce \textbf{\textsc{NoTeS-Bank}}, an \textit{evaluation benchmark} for \textbf{Neural Transcription and Search} in note-based question answering. \textsc{NoTeS-Bank} comprises complex notes across multiple domains, requiring models to process unstructured and multimodal content. The benchmark defines two tasks: (1) \textbf{Evidence-Based VQA}, where models retrieve localized answers with bounding-box evidence, and (2) \textbf{Open-Domain VQA}, where models classify the domain before retrieving relevant documents and answers. Unlike classical Document VQA datasets relying on optical character recognition (OCR) and structured data, \textsc{NoTeS-BANK} demands vision-language fusion, retrieval, and multimodal reasoning. We benchmark state-of-the-art Vision-Language Models (VLMs) and retrieval frameworks, exposing structured transcription and reasoning limitations. \textsc{NoTeS-Bank} provides a rigorous evaluation with \textit{NDCG@5, MRR, Recall@K, IoU, and ANLS}, establishing a new standard for visual document understanding and reasoning. 
\end{abstract}




\maketitle

\begin{figure*}[ht]
    \centering
    \includegraphics[width=\textwidth]{motivation_teaser_NoTeS.pdf}
    \caption{\textit{Comparison of OCRs on a challenging handwritten scientific note sample.} Powerful commercial OCR engines (eg. Textract, Google-OCR) fail to accurately transcribe the content, often losing mathematical symbols, structure, and semantic meaning. In contrast, open-source OCRs (e.g. Nougat, Got 2.0, OLM-OCR) struggle to extract even the textual content for the full document. This highlights the \textit{necessity for multimodal reasoning beyond OCR} in handwritten document understanding.}
    \label{fig:sankey}
\end{figure*}

\section{Introduction}

\textit{“What we know is a drop, what we do not know is an ocean.”} – Isaac Newton. The pursuit of knowledge often begins with handwritten notes — scribbled equations, diagrams, and annotations that serve as the foundation of scientific discovery, engineering breakthroughs, and academic learning. However, despite the fundamental role of handwritten notes, their automated understanding remains a formidable challenge in visual document understanding (VDU). Prior datasets~\cite{mathew2021asking} cover either neatly rendered handwriting (HW-SQuAD) or historical letters (BenthamQA), but none address modern lecture or notebook pages with complex content. Academic notes often mix prose with formulas, contain shorthand or abbreviations, and include sketches, flowcharts, or diagrams drawn by the note-taker. Current state-of-the-art (SOTA) OCR-based VDU models~\cite{xu2020layoutlm,huang2022layoutlmv3,appalaraju2021docformer} assume a well-defined layout or reading order as in DocVQA~\cite{mathew2020document}. They can falter when content is scattered or non-linear. For instance, models in such datasets~\cite{mathew2020document, van2023document} performed poorly on questions where the answer required interpreting the document’s structure (tables, columns, or aligned layout elements). In handwritten lecture notes, the “layout” might include diagrams next to text or equations below explanatory text, which is not trivial for models to parse, as shown in Figure~\ref{fig:sankey}.

\begin{figure}
    \centering
    \includegraphics[width=\linewidth]{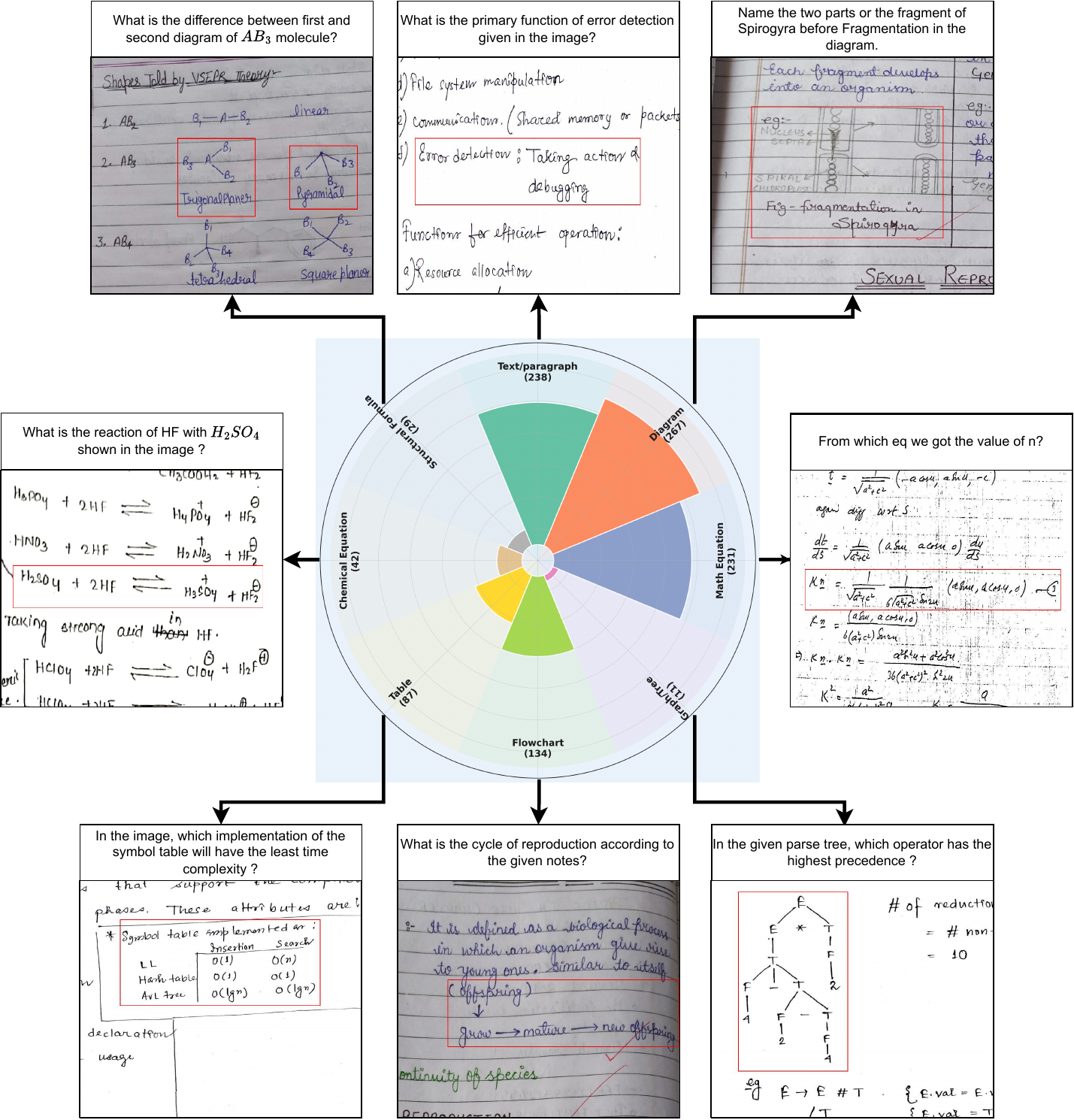}
    \caption{\textit{Illustrative examples from the \textsc{NoTeS-Bank}} showing diverse local reasoning categories such as equations, flowcharts, structural formulas, and textual answers. Center: Distribution of question types across eight task-level categories. Surrounding: Sample questions and annotated evidence regions from handwritten notes, highlighting the visual and semantic complexity handled in the Evidence-Based VQA task.}
    \label{fig:NoTeS_bank}
\end{figure}

Most existing models are not equipped to jointly analyze visual drawings and text. They either ignore non-text elements or treat the problem as pure text extraction or OCR tasks. This is a limitation when notes contain sketches, scientific diagrams, or mathematical equations. For example, a flowchart or a geometric diagram might be essential to answer a question, but generic text-based VQA models will not interpret a drawing of a triangle or a circuit diagram. Similarly, handwritten equations or symbols (e.g. $\Sigma$, integrals, chemical structures) can be misread by OCR or not understood in context (e.g., distinguishing a handwritten “z” from a “2” in an equation as shown in Figure~\ref{fig:sankey}). Existing benchmarks like InfographicVQA~\cite{mathew2022infographicvqa} and MathVista~\cite{lu2023mathvista} explicitly show that joint reasoning over text and graphics is needed, and current SOTA baselines perform modestly on such tasks. In summary, today’s VQA systems tend to be brittle outside the domains they were trained in - they struggle with messy handwriting, unstructured layouts, and mixed modalities prevalent in note-based documents. The information is there—somewhere—but buried beneath uneven ink, equations in the margins, half-drawn diagrams, and scribbled corrections. To answer a question, they don’t just recall the fact—they search, they scan, they locate the exact line, the boxed formula, the circled term. That process of not only knowing the answer but being able to localize to it is at the heart of our \textbf{evidence-based VQA} task. Evidence-based document VQA challenges models to look at the document like a human would, navigating through visual cues, spatial arrangements, and noisy handwriting to retrieve and ground the answer as shown in Figure~\ref{fig:NoTeS_bank}. This isn’t just about accuracy—it’s about understanding. Grounding the response builds trust, explains reasoning, and proves that the model isn’t just parroting patterns but truly reading the document. In our benchmark, this grounding becomes the litmus test: \textit{can your model not only answer the question but also show its work?} The key challenge is to answer questions by identifying the correct region inside a visual academic note (via bounding box), while also classifying both the local content type (e.g. flowchart, chemical formulae, graph, table, math equation, etc.) and the global domain (e.g., physics, computer science) of the answer—encouraging a full spectrum of document understanding, from retrieval to reasoning.

Now, picture a student preparing for an oral exam, trying to answer a complex, open-ended question like, "\textit{Why does underflow occur in floating point representation?}" The answer probably lives somewhere in a series of lecture notes, in different subjects, on different days. To respond, the student first narrows it down: this sounds like computer science or numerical methods. Then, they dive into their notes, flipping through the relevant sections and scanning for definitions, diagrams, or formulas. Only then do they piece together an answer—sometimes from multiple fragments, maybe with a margin sketch or an annotated equation to help. This is the essence of \textbf{open-domain question answering} over academic notes. But unlike vanilla open-domain QA systems~\cite{marino2019ok} or M3DocVQA~\cite{cho2024m3docrag} that pull structured facts from curated sources like Wikipedia, our task demands that the model acts like a student, knowing where to look, what to extract, and how to connect the dots visually and semantically. The challenge isn't just to answer but to find a filter and reason across noisy, visual content.
In our setup, models must do three things: first, predict the domain (e.g., physics, mathematics, chemistry) from the question to guide retrieval; second, retrieve relevant pages of handwritten notes—not structured web articles or knowledge bases; and third, perform multimodal reasoning across diagrams, equations, and prose, often scattered across the page. 

\begin{figure}
    \centering
    \includegraphics[width=0.6\linewidth,height=8cm]{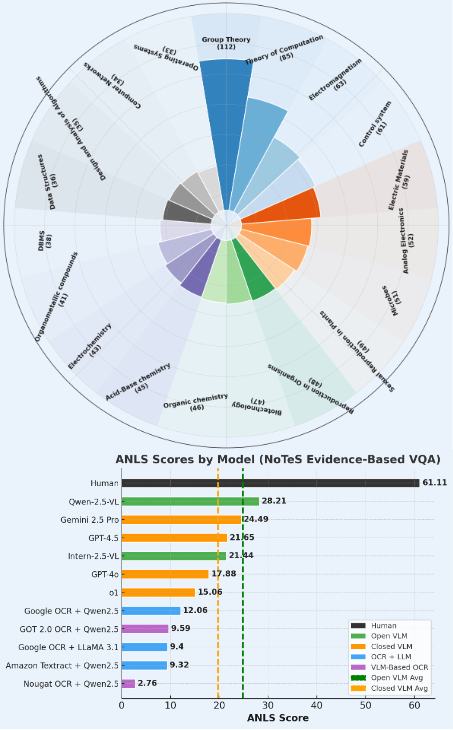}
    \caption{(Top) \textit{Global category distribution} across 19 scientific and technical domains in the NoTeS-Bank benchmark, including physics, biology, chemistry, and computer science. This diverse coverage enables a fine-grained evaluation of domain-specific reasoning. (Bottom) ANLS performance comparison on the Evidence-Based VQA task. Results are shown for human annotators, open/closed Vision-Language Models (VLMs), and OCR+LLM pipelines. The large performance gap highlights the challenge of accurately answering and grounding questions in visually unstructured, handwritten academic notes.}
    \label{fig:NoTeS_bank_cat}
\end{figure}

 Though significant advancements have been witnessed in recent times with the emergence of Vision-Language Models (VLMs)~\cite{laurenccon2024matters,wang2024qwen2} and retrieval-augmented approaches~\cite{cho2024m3docrag}, note-based document VQA remains a challenging, underexplored problem. 
 Unlike structured document databases~\cite{zhong2019publaynet, pfitzmann2022doclaynet}, scientific notes exhibit \textit{unstructured layouts, informal writing styles, and a lack of clear segmentation}, making traditional OCR-based approaches unreliable. Furthermore, questions often require \textbf{multimodal reasoning}: understanding mathematical derivations, interpreting scientific notation, or localizing diagrams that visually complement textual explanations. VLMs often excel by picking up on textual cues, but for notes, visual context (like an arrow connecting a note to a figure or text style like underlining) can change the meaning. Many models do not fully leverage these cues. OCR-based pipelines~\cite{blecher2023nougat, wei2024general} “throw away” visual information by converting image to text, thus losing layout, handwriting style, or markings as shown in Figure~\ref{fig:sankey}. As one recent work~\cite{faysse2024colpali} states, modern document retrieval systems mainly rely on extracted text and miss “key visual cues” like figures or layout, limiting their capability on complex documents. This limitation underscores the need for approaches that \textit{treat the image of the document as a first-class input, not just the transcribed text.} In essence, current VQA or VDU models have a \textbf{modality gap} – they handle printed text well, but performance drops on unstructured, handwritten, or multi-modal inputs. This is exactly the challenge \textsc{NoTeS-BANK} is designed to address. While most document understanding datasets focus on either layout segmentation or holistic document classification, our benchmark introduces a \textit{dual-layer annotation schema} that grounds each QA instance both at the local (task) level—such as interpreting equations, diagrams, or tables—and the global (domain) level, encompassing scientific fields like physics, biology, or computer science as shown in Figure~\ref{fig:NoTeS_bank_cat}. This fine-grained semantic labeling enables deeper insight into model behavior across multimodal reasoning types and subject domains, making \textsc{NoTeS-Bank} not only a QA benchmark but also a diagnostic tool for analyzing the scope and limits of VLMs and multimodal RAGs. The key contributions of this work can be summarized as follows: 1) We present \textsc{NoTeS-Bank}, a novel benchmark for question answering over unstructured, scientific notes, addressing a gap in multimodal document understanding by focusing on visio-graphical content beyond printed or structured formats. 2) We define two tasks—Evidence-Based VQA and Open-Domain VQA—that jointly evaluate answer grounding, domain classification, and retrieval-based multimodal reasoning in challenging handwritten scenarios. 3) We benchmark a diverse set of VLMs, OCR+LLM pipelines, and retrieval-augmented approaches, and provide a comprehensive evaluation framework using ANLS, IoU, Recall@K, and MRR to highlight the modality gap in current systems.

\begin{table*}[t]
\caption{Comparison of benchmarks on content type, document setting, domain, and task type.}
\label{tab:benchmark-comparison}
\begin{tabularx}{\textwidth}{@{}lL{3.2cm}cL{2.5cm}Y@{}}
\toprule
\textbf{Benchmark} & \textbf{Content Type} & \textbf{Multi Document} & \textbf{Domain} & \textbf{Tasks} \\ 
\midrule
LongBench~\cite{bai2023longbench} & Text & \cmark & Wikipedia & Long-form QA, Retrieval \\
MPDocVQA~\cite{tito2022hierarchical} & Text, Tables, Charts & \xmark & Multi-domain & Document Visual QA \\
$\infty$Bench~\cite{zhang2024infty} & Text & \xmark & Multi-domain & List QA, Reasoning \\
DUDE~\cite{van2023document} & Text, Tables, Charts, Figures & \xmark & Multi-domain & Document Visual QA, Muti-hop QA, Unanswerable, List QA \\
MMLONGBENCH-DOC~\cite{ma2024mmlongbench} & Text, Tables, Charts, Slides & \xmark & Multi-domain & Document QA, List QA \\
M3DocVQA~\cite{cho2024m3docrag} & Text, Tables, Charts & \textbf{\cmark} & Wikipedia & Open-domain Document VQA \\
VisDoMBench~\cite{suri2024visdom} & Text, Tables, Charts, Slides & \textbf{\cmark} & Multi-domain & Evidence-based Visual Grounding with Bbox, Open-domain QA \\
\midrule
\textbf{\textsc{NoTeS-Bank} (Ours)} & \textbf{Graphical Diagram, Math Equation, Chemical Equation, Structural formula, Text/paragraph, Graph/Tree, Flowchart, Table} & \textbf{\cmark} & \textbf{Multi-domain (Scientific), Multi-Task} & \textbf{Evidence-based Visual Grounding with Bbox and Semantic Labeling, Open-domain VQA, Multi-hop QA, Unanswerable QA, Reasoning} \\
\bottomrule
\end{tabularx}
\label{tab:related}
\end{table*}

\section{Related work}

VQA provides a natural language interface for tackling diverse vision-language tasks, merging computer vision and natural language processing (NLP) techniques. This approach has been widely applied across multiple domains, including medical question answering \cite{Nentidis_2022, jin-etal-2019-pubmedqa, raghavan-etal-2021-emrkbqa}, open-domain knowledge retrieval \cite{yang-etal-2015-wikiqa, https://doi.org/10.48550/arxiv.2007.15207, https://doi.org/10.48550/arxiv.2004.10645, liu-etal-2019-xqa}, emotion recognition \cite{gui-etal-2017-question, bjerva-etal-2020-subjqa}, code-based QA \cite{agashe-etal-2019-juice, liu-wan-2021-codeqa-question}, logical reasoning \cite{ijcai2020p0501, yu2020reclor}, fact verification \cite{hu-etal-2022-chef, zarharan-etal-2021-parsfever}, and mathematical reasoning \cite{zhang-etal-2021-noahqa-numerical, hopkins-etal-2019-semeval, chen-etal-2021-geoqa, lu2023mathvista}. 

The field of VDU has been fueled by new benchmarks that invoke a multimodal understanding of Document VQA models. Earlier datasets like DocVQA~\cite{mathew2020document,tito2021document} and InfographicsVQA~\cite{mathew2022infographicvqa} evaluated reading comprehension on single pages or images, but recent benchmarks cover more diverse and realistic scenarios. Notably, the Document Understanding Dataset and Evaluation (DUDE)~\cite{dude2023icdar,van2023document} is a large-scale multi-page, multi-domain DocVQA benchmark which spanned documents from many industries and layouts, with questions requiring reasoning across lengthy documents. It reported that even SOTA models (layout-aware Transformers~\cite{xu2020layoutlm,huang2022layoutlmv3} and VLMs~\cite{li2022blip,li2023blip,alayrac2022flamingo,achiam2023gpt,chattopadhyay2024towards}) perform far below human accuracy in DUDE, highlighting the difficulty of generalizing across domains~\cite{das2024diving}. Another benchmark, SlideVQA~\cite{tanaka2023slidevqa}, focuses on presentation slides with complex layouts, while TableVQA~\cite{kim2024tablevqa} and ScreenUI~\cite{baechler2024screenai} tasks have introduced questions requiring understanding tables or user interface-like documents. To evaluate the understanding of long documents, MMLongBench-Doc~\cite{ma2024mmlongbench} was proposed as part of a long-context multimodal suite featuring approximately 50-page scientific articles and reports to test how models handle complex sequential layouts. In the open-source community, MMDocBench~\cite{zhu2024mmdocbench} compiled 15 diverse document tasks (from receipts and research papers to diagrams) with over 4,000 QA pairs to compare various large VLMs in a zero-shot setting.  This benchmark revealed strengths and weaknesses of models like GPT-4V~\cite{achiam2023gpt}, LLaVA~\cite{liu2023visual}, and InternVL~\cite{chen2024expanding} on fine-grained document tasks, and it provides a comprehensive testbed for OCR-free document understanding across formats. Finally, new datasets are coupling document images with open-domain knowledge: for example, VisDomRAG~\cite{suri2024visdom} evaluates QA systems in multi-document settings with rich multimodal content (tables, charts, and presentation slides) while M3DocVQA~\cite{cho2024m3docrag} does it for enterprise documents for testing multimodal Retrieval Augmented Generation (RAG) approaches. The surge of these benchmarks - from DUDE's industry-spanning documents to M3DocVQA's open domain corpus (as shown in Table~\ref{tab:related}) is driving the field toward more robust and generalizable document understanding, ensuring that modern VLMs are evaluated on reading, reasoning, and retrieving information from the full variety of documents encountered in the wild. Building on this momentum, our work takes a significant step further by introducing \textsc{NoTeS-Bank}, \textit{the first benchmark focused entirely on handwritten, purely visual, and unstructured academic notes} — challenging current models to reason without structured text anchors or reliable OCR, and establishing a new frontier for document understanding in the wild.

\begin{figure*}[t]
    \centering
    \includegraphics[width=\textwidth, height= 8.7cm]{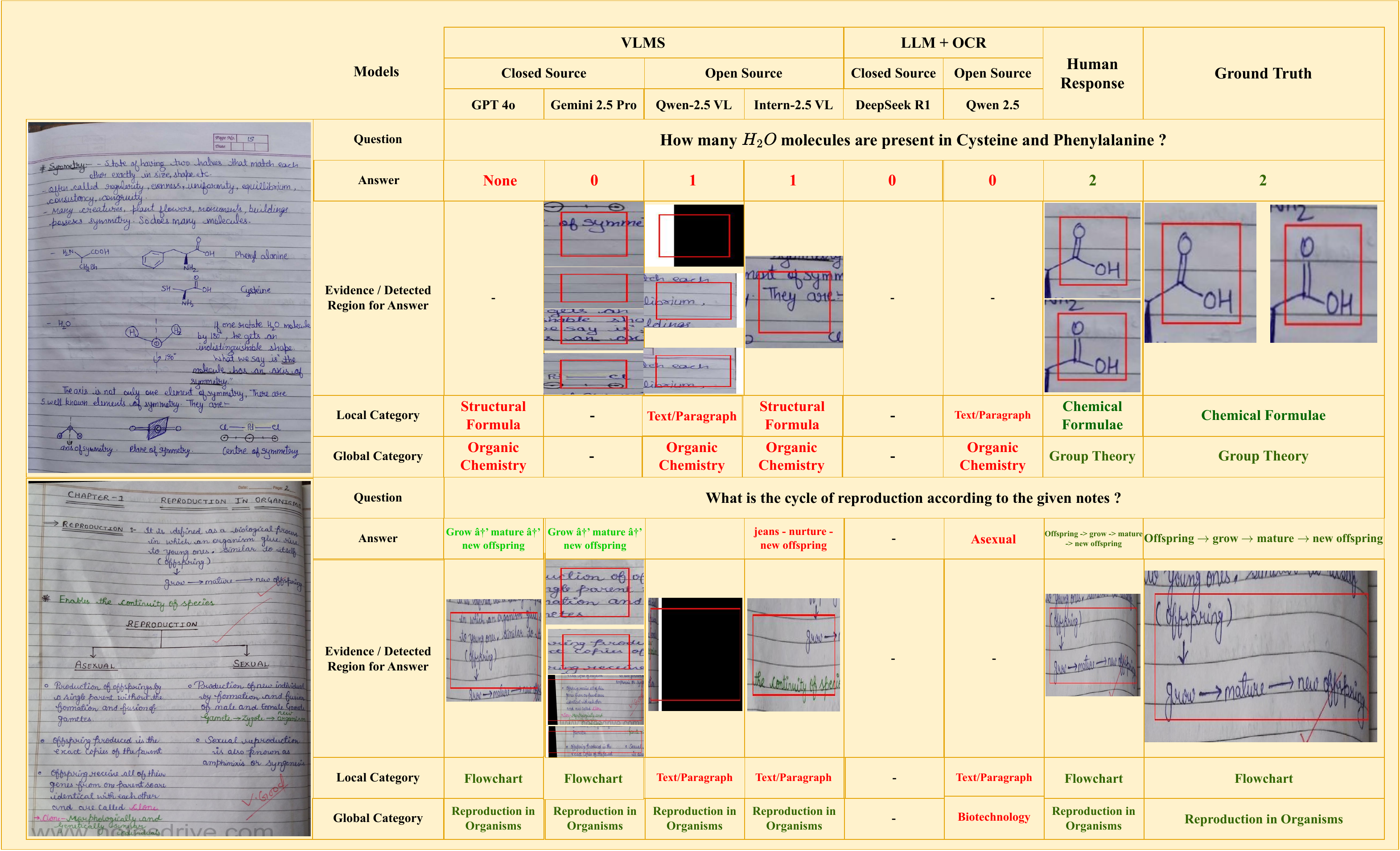}
    \caption{\textit{Qualitative comparison of Vision-Language Models (VLMs), OCR+LLMs, and human responses on the NoTeS-Bank Evidence-Based VQA task}. Each example demonstrates the challenge of retrieving grounded answers from handwritten scientific notes, highlighting the limitations of current models in detecting accurate regions and reasoning over domain-specific content. The figure also illustrates the fine-grained local (e.g., Structural Formula, Flowchart) and global (e.g., Organic Chemistry, Reproduction in Organisms) category annotations provided for each question-answer pair.}
    \label{fig:enter-label}
\end{figure*}

\section{The \textsc{NoTeS-Bank} Benchmark Suite}

\textsc{NoTeS-Bank} is a gold-standard evaluation benchmark designed to assess multimodal question answering over complex unstructured scientific notes. Unlike existing DocVQA datasets~\cite{mathew2020document,tito2021document,mathew2022infographicvqa,van2023document,mathew2021asking}, \textsc{NoTeS-BANK} introduces two distinct tasks that challenge vision-language models (VLMs) and multimodal retrieval-augmented generative (RAG) architectures.

\subsection{Evidence-Based VQA} The Evidence-Based VQA (EB-VQA) task in \textsc{NoTeS-BANK} evaluates a model’s ability to retrieve, comprehend, and justify answers using handwritten note-based evidence. Unlike existing DocVQA challenges that rely solely on extracted OCR text, this task requires models to reason over visual semantics, structural elements, and handwritten symbols while ensuring explicit grounding of responses.\\

\noindent
\textbf{Task Formulation:} Given an input visual note (image) \( I \) (which could span 1-3 pages) containing unstructured text, symbols, equations and diagrams, and a natural language question \( Q \), the model must: (a) \textit{Retrieve Relevant Evidence}: Identify key portions of the visual note \( I \) that contribute to answering \( Q \) \textit{by means of a bounding box or multiple bounding boxes}. (b) \textit{Generate an Answer}: Synthesize a natural language response \( A \) based on the retrieved evidence \( E \). (c) \textit{Provide Justification}: Highlight the supporting evidence \( E \) in \( I \) that links to the final answer, including the corresponding visual elements (e.g., mathematical equations, chemical formulas, diagrams).

Formally, the model is defined as:

\begin{equation}
    A, E = f_{\text{EB-QA}}(I, Q)
\end{equation}

where the evidence set \( E \) consists of:
\begin{equation}
    E = \{ (B_i, L_i, G_i) \}_{i=1}^{p}
\end{equation}
where:
- \( B_i \) represents the bounding box of the relevant evidence region,
- \( L_i \) denotes the local category of the evidence (e.g., equation, table, diagram),
- \( G_i \) specifies the global category related to the document's conceptual domain (e.g., group theory, rotational mechanics).\\

\noindent
\textbf{Evaluation:}
To evaluate model performance, we assess answer accuracy and evidence selection quality. Answer accuracy is measured using Average Normalized Levenshtein Similarity (ANLS), while evidence selection is evaluated through Intersection-over-Union (IoU), which quantifies alignment between predicted evidence \( E \) and the ground truth \( E^* \). 

\begin{equation}
    IoU(E, E^*) = \frac{|E \cap E^*|}{|E \cup E^*|}
\end{equation}

Additionally, we measure the correctness of local and global element categorization inside, ensuring that the models retrieve not only the appropriate text regions but also the relevant semantic concepts necessary for reasoning.

\subsection{Open-Domain Question Answering}

The Open-Domain QA (OD-QA) task in \textsc{NoTeS-BANK} evaluates a model’s ability to retrieve, reason, and generate answers across a large collection of handwritten notes. Unlike standard document QA tasks that operate within a single document, this task requires models to first classify the domain of the question, retrieve the most relevant handwritten document, and then generate an answer.

Given a document collection \( D \) and a natural language question \( Q \), the model must predict the subject category \( C \), retrieve the most relevant document \( I \), and generate the final answer \( A \):

\begin{equation}
    C = f_{\text{domain}}(Q),  \quad I = f_{\text{retrieve}}(D, Q, C), \\
    \quad A = f_{\text{answer}}(I, Q)
\end{equation}

where:
- \( C \) represents the predicted subject category (e.g., physics, mathematics).
- \( I \) is the retrieved handwritten document.
- \( A \) is the generated answer.

Unlike traditional retrieval-based QA, \textsc{NoTeS-BANK} requires models to handle noisy, unstructured, and multimodal content, including mathematical expressions, diagrams, and scientific notations. The retrieval component is evaluated using Hit@K, MRR, and NDCG@5 to assess ranking quality, while answer accuracy is measured through ANLS. Additionally, the correctness of document and page selection is assessed to ensure models retrieve and process the most relevant content.

\begin{figure}[h]
    \centering
\includegraphics[width=0.8\linewidth]{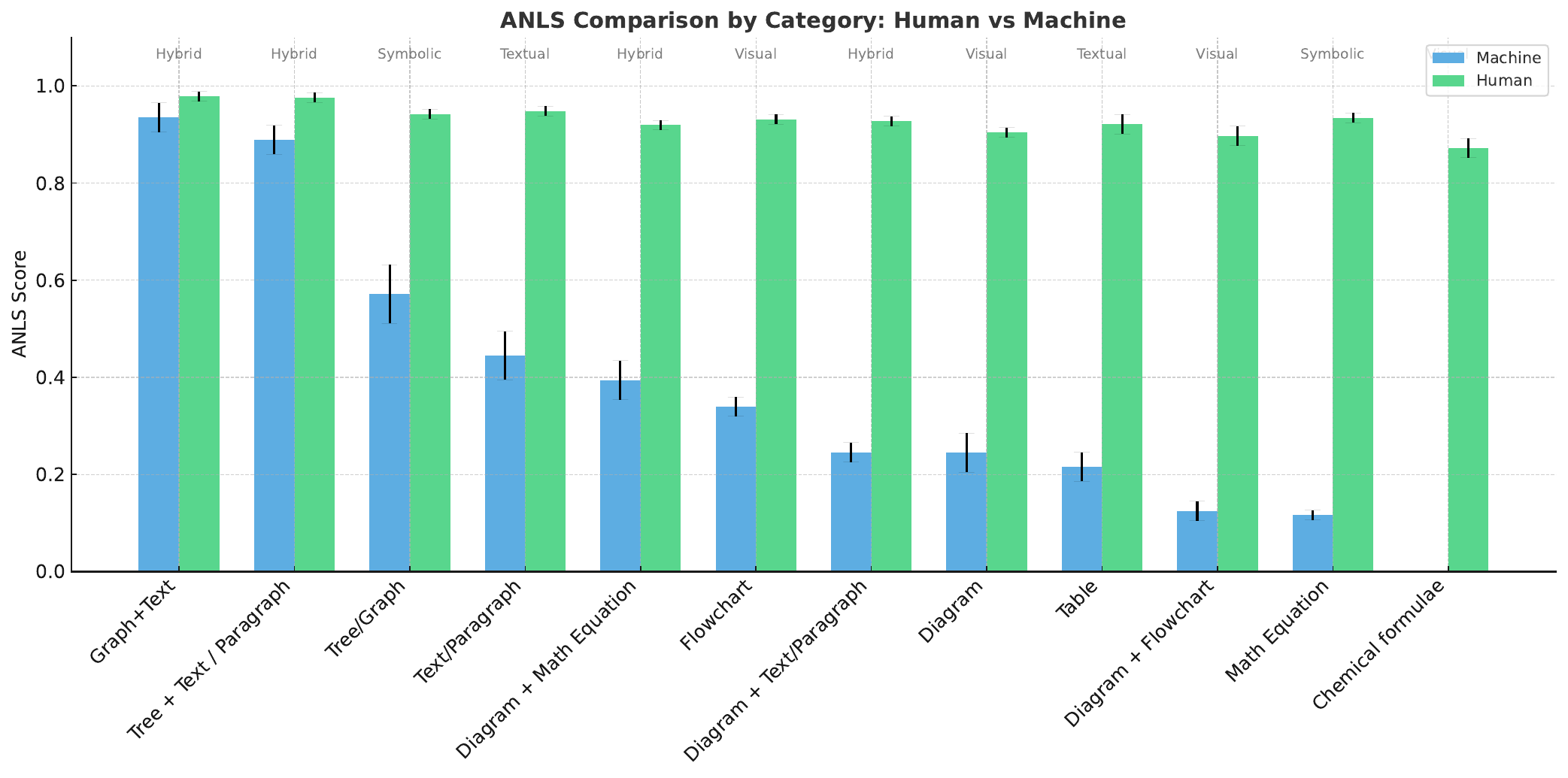}
    \caption{We report the average ANLS for the human expert vs. the best-performing model per diagnostic category as a ceiling analysis.} 
    \label{fig:}
\end{figure}

\subsection{Dataset Collection and Annotation}

\noindent
\textbf{Data Collection.}
The dataset was collected from various educational websites and student study materials, including resources such as Gate Overflow and self-study notes. For websites where the materials were hosted for students, explicit permission was obtained from the respective content owners, who consented to the use of their data for research purposes. The dataset includes a diverse range of handwritten notes covering subjects such as physics, mathematics, chemistry, biology, and engineering.

\vspace{0.1cm}
\noindent
\textbf{Annotation Process.}
A team of 15–20 annotators, primarily undergrad students, was assembled to annotate the dataset. Multiple training sessions were conducted before starting the annotation process to ensure consistency in question formulation and evidence selection. These sessions familiarized annotators with the criteria for both tasks and helped standardize the approach across different document types.

\vspace{0.1cm}
\noindent
\textbf{Task 1: Evidence-Based QA Annotation.} For the Evidence-Based QA task, annotators first created question-answer pairs by formulating queries that required reasoning over textual and multimodal content. Each QA pair was recorded along with metadata, including whether the document was single-page or multi-page, the page number where the answer appeared, and the subject name. To further categorize the nature of the answer, the region from which the answer was derived was labeled as a local category, such as text, equation, diagram, or chemical formula. Each QA pair was also assigned a global category corresponding to its conceptual domain, such as group theory or rotational mechanics.

Once the QA pairs were created, annotators manually labeled the corresponding answer regions by drawing bounding boxes around the relevant content using an annotation tool. These bounding boxes, along with the associated QA pairs and document images, were compiled into a structured JSON format for experimentation and evaluation.

\vspace{0.1cm}
\noindent
\textbf{Task 2: Open-Domain QA Annotation.} A separate annotation process was conducted for the Open-Domain QA task to ensure that retrieval-based reasoning was accurately represented. A new set of QA pairs was created to require retrieval across multiple documents rather than direct extraction from a single page. In this task, the domain classification of each QA pair was explicitly recorded to assist in retrieval, ensuring that models could infer the subject category before searching for the answer. Additionally, annotators identified the ground truth document and the specific page from which the answer should be retrieved. Given the fundamental differences between the two tasks, the dataset was annotated in two independent rounds to maintain consistency and avoid overlap in annotation strategies. This approach ensured that models trained on the dataset would be evaluated on both localized, evidence-grounded reasoning and retrieval-based question answering across an open-domain handwritten note collection.

\begin{table*}[t]
    \centering
    \scriptsize
     \caption{Performance comparison of Open and Closed VLM-Based models, OCR + LLM, Layout + OCR + LLM, and VLM-Based OCR models on the Evidence-Based VQA task in \textsc{NoTeS-BANK}.RL: Region-Level Layout; WL: Word-Level Layout}
    \begin{tabular}{lcc|c|ccc|cc}
        \toprule
        \textbf{Model} & \textbf{\#Param} & \textbf{Context Window} & \textbf{ANLS*} & \multicolumn{3}{c}{\textbf{IoU Metrics}} & \multicolumn{2}{c}{\textbf{Category Accuracy (\%)}} \\
        & & & & Avg IoU & IoU@5 & IoU@10 & Local & Global \\
        \midrule
        \multicolumn{9}{l}{\textit{Open VLM-Based Models}}\\
        Qwen-2.5-VL~\cite{wang2024qwen2} & 7B & 32K & 28.21 & 0.0136 & 0.0727 & 0.0518 & 11.83 & 4.86 \\
        Intern-2.5-VL~\cite{chen2024expanding} & 8B & 16k & 21.44 & 0.0097 & 0.049 & 0.0308 & 6.47 & 7.91 \\
        LLaVA-OneVision~\cite{li2024llava} & 8B & 60k & 23.34 & - & - & - & - & - \\
        \midrule
        \multicolumn{9}{l}{\textit{Closed VLM-Based Models}}\\
        GPT-4o~\cite{achiam2023gpt} & - & 128k & 17.88 & 0.0122 & 0.036 & 0.0381 & 12.0 & 9 \\
        Gemini 2.5 Pro~\cite{team2023gemini} & - & 2M & 24.493 & 0.013 & 0.016 & 0.0367 & 0.2 & - \\
        OpenAI o1~\cite{openai2024openaio1card} & - & 100k & 15.06 & 0.0107 & 0.0565 & 0.0335 & 15.6 & 11.2\\
        GPT-4.5~\cite{achiam2023gpt} & - & 128k & 21.65 & 0.0186 & 0.0898 & 0.0579 & 13.4 & 7.19 \\
        \midrule
        \multicolumn{9}{l}{\textit{OCR + Closed LLM Models}}\\
        Google OCR~\cite{hegghammer2022ocr} + Deepseek-R1~\cite{liu2024deepseek} & 7B & 128k & 12.46 & - &- & -& 1.04 & 0.2283 \\
        \midrule
        \multicolumn{9}{l}{\textit{OCR + Open LLM Models}}\\
        Google OCR~\cite{hegghammer2022ocr} + Qwen2.5~\cite{wang2024qwen2} & 7B & - & 12.06 & 0& 0 & 0 & 7.6 & 4.8 \\
        Amazon Textract~\cite{hegghammer2022ocr} + Qwen2.5~\cite{wang2024qwen2} & 7B & - & 9.32 & 0 & 0 & 0 & 9.07 & 0.824 \\
        Google OCR~\cite{hegghammer2022ocr} + LLaMA 3.1~\cite{dubey2024llama} & 8B & - & 9.4 & 0.0077 & 0.04865 & 0.02001 & 5.78 & 3.21 \\
        Amazon Textract~\cite{hegghammer2022ocr}  + LLaMA 3.1~\cite{dubey2024llama} & 8B & - & 7.23 & 0.0032 & 0.0189 & 0.0101 & - & - \\
        Amazon Textract~\cite{hegghammer2022ocr} + LLaMA 3.1~\cite{dubey2024llama} + RL & 8B & - & 5.37 & 0 & 0 & 0 & 0.4124 & 0.6185 \\
        Amazon Textract~\cite{hegghammer2022ocr} + LLaMA 3.1~\cite{dubey2024llama} + WL & 8B & - & 7.54 & 0.0042 & 0.0261 & 0.0132 & 0.2062 & 1.6494 \\
        Amazon Textract~\cite{hegghammer2022ocr} + LLaMA 3.1~\cite{dubey2024llama} + RL + WL & 8B & - & 7.01 & 0.0032 & 0.0065 & 0.0065 & 0.2061 & 1.4432 \\
        
        \midrule

        \multicolumn{9}{l}{\textit{VLM-Based OCR Models}}\\
        Nougat OCR~\cite{blecher2023nougat} + LLaMA 3.1~\cite{dubey2024llama} & - & - & 3.2 & - & - & - & ~0 & ~0 \\
        GOT 2.0 OCR~\cite{wei2024general} + LLaMA 3.1~\cite{dubey2024llama} & - & - & 1.73 & - & - & - & ~0 & ~0 \\
        olmOCR~\cite{poznanski2025olmocr} + LLaMA 3.1~\cite{dubey2024llama} & - & - & 4.86 & 0.0001 & 0.0011 & ~0 & ~0 & ~0 \\
        Nougat OCR~\cite{blecher2023nougat} + Qwen2.5~\cite{wang2024qwen2} & - & - & 2.76 & - & - & - & 3.8 & 1.2 \\
        GOT 2.0 OCR~\cite{wei2024general} + Qwen2.5~\cite{wang2024qwen2} & - & - & 9.59 & - & - & - & 3.4 & 0.8 \\
        olmOCR~\cite{poznanski2025olmocr} + Qwen2.5~\cite{wang2024qwen2} & - & - & 4.69 & - & - & - & 2.6 & 1.6 \\
        \midrule
        \textbf{Human Baseline} & - & - & 61.11 & 0.4009 & 0.5 & 0.4312 & 83 & 79 \\
        \bottomrule
    \end{tabular}

    \label{tab:evidence_based_vqa}
\end{table*}

\subsection{Baselines and Model Selection}

To evaluate performance on the Evidence-Based VQA task, we establish diverse baselines covering vision-language models (VLMs), OCR-based pipelines, and retrieval-augmented approaches. These baselines assess how different model architectures handle handwritten documents, reasoning, and evidence selection.

\vspace{0.1cm}
\noindent
\textbf{Vision-Language Models (VLMs).} VLMs holistically process visual and textual features, making them well-suited for handwritten notes where OCR struggles. We benchmark both \textit{open} (Qwen-2.5-VL, Intern-2.5-VL, LLaVA) and \textit{closed} (GPT-4o, Gemini 2.5-Pro, GPT-4.5, O1) models to analyze their multimodal reasoning capabilities.

\vspace{0.1cm}
\noindent
\textbf{OCR + LLM-Based Models.} Traditional OCR pipelines extract text before reasoning, but handwritten documents introduce recognition errors. We evaluate Google OCR, Amazon Textract, and OLM OCR combined with LLaMA 3.1 to measure OCR impact on answer quality.

\vspace{0.1cm}
\noindent
\textbf{Layout + OCR + LLM Models.} Standard OCR pipelines discard document structure. We introduce layout-aware approaches incorporating region- and word-level cues (e.g., Textract + Layout Prompt) to assess document retrieval and reasoning improvements.

\vspace{0.1cm}
\noindent
\textbf{VLM-Based OCR Models.} Instead of explicit text extraction, models such as Nougat OCR and GOT 2.0 OCR attempt direct transcription using vision-language understanding. These baselines highlight the limitations of OCR-free approaches in handwriting recognition. Models are compared across \textit{ANLS} (answer similarity), \textit{IoU} (evidence selection), and \textit{category accuracy} (domain classification) to capture end-to-end document understanding. By selecting these baselines, we establish a comprehensive benchmark to drive advancements in handwritten document QA.

\vspace{0.1cm}
\noindent
\textbf{Human performance.} For human evaluation, we collected responses from domain-aware individuals who were not involved in the dataset annotation process. Each participant independently answered questions and highlighted the corresponding evidence regions within the handwritten documents. This separation ensured unbiased assessment across both tasks. Human responses consistently outperformed automated models in terms of accuracy and grounding, especially for complex, multimodal queries—highlighting the difficulty of the \textsc{NoTeS-Bank} benchmark and the gap between current model capabilities and expert-level understanding.


\begin{figure*}
    \centering
    \includegraphics[width=1\linewidth, height= 9cm]{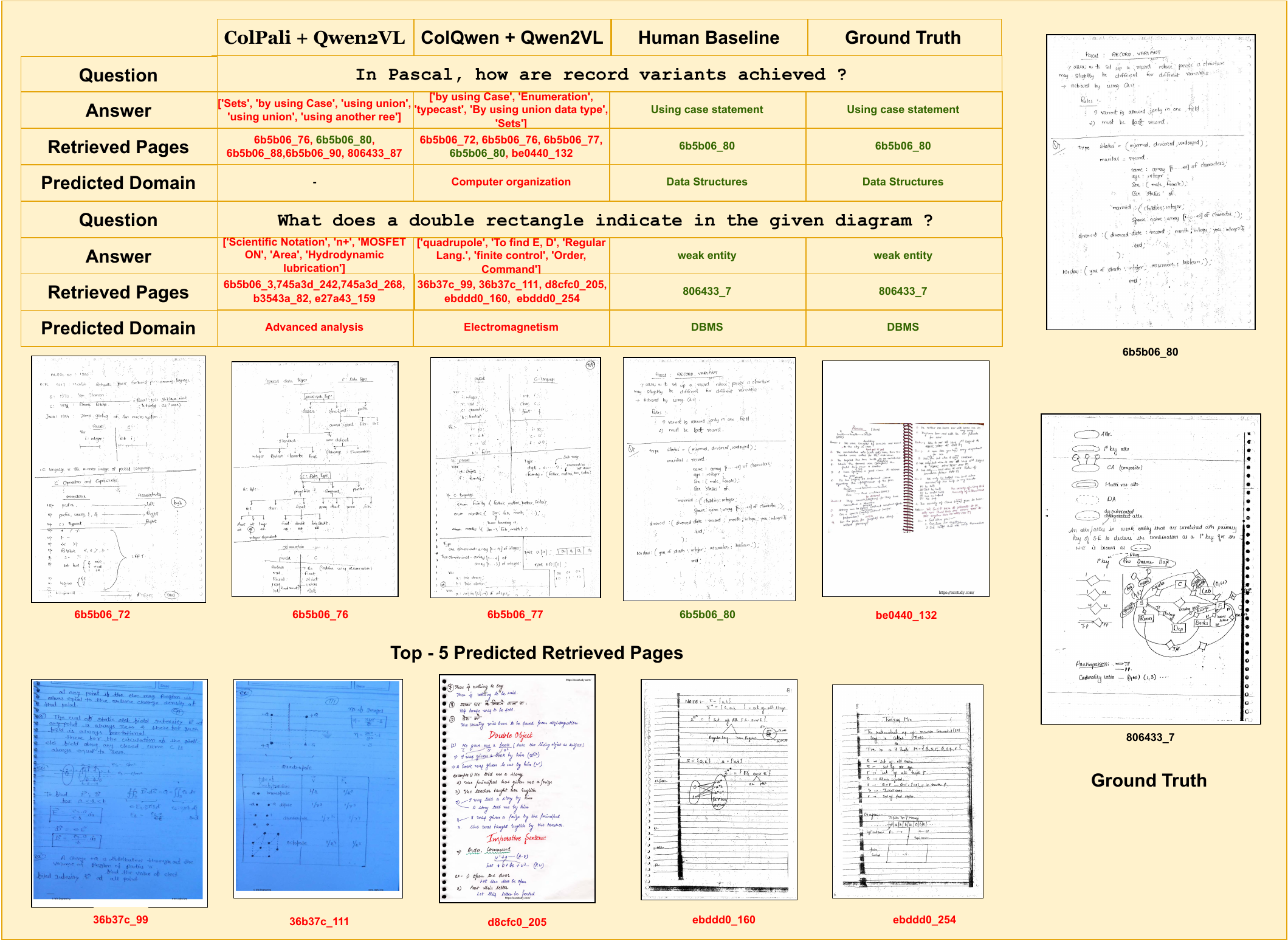}
    \caption{\textit{Qualitative comparison of Open-Domain VQA performance in the NoTeS-Bank benchmark.} Each query requires retrieving relevant handwritten pages from a large corpus and reasoning across them to answer the question. The figure highlights model predictions from retrieval-augmented VLMs (ColPali + Qwen2VL and ColQwen + Qwen2VL) alongside human responses and ground-truth annotations. Differences in predicted answers, retrieved documents, and domain classification underscore the challenge of joint retrieval, domain inference, and multimodal reasoning over noisy, unstructured visual content.}
    \label{fig:enter-label}
\end{figure*}

\section{Results and Discussion}
\label{sec:results}

\subsection{Evaluation Protocol}

Prior work~\cite{yang2023dawn} has explored the reasoning capabilities of foundation models in visual tasks, primarily through qualitative analysis. In contrast, our objective with \textsc{NoTeS-BANK} is to establish a systematic and unified evaluation protocol that enables both quantitative and qualitative assessment of vision-language models for symbolic and multimodal reasoning in handwritten scientific documents. We introduce a comprehensive benchmarking strategy for \textsc{NoTeS-BANK}, encompassing both Evidence-Based VQA and Open-Domain QA). The models included in our benchmark range from OCR-enhanced LLMs to open and closed-source vision-language models, as detailed in Table~\ref{tab:qa-retrieval}. We report results across multiple evaluation dimensions, including answer correctness (ANLS*), evidence localization (IoU), document retrieval (Recall@K, MRR, NDCG@5), and category prediction accuracy in Sec.~\ref{sec:results}.

In addition to quantitative performance metrics, we also provide qualitative analysis of representative failure cases and model outputs, shedding light on limitations in layout reasoning, symbol understanding, and evidence attribution. Given the relatively stronger performance of GPT-4o in multimodal tasks, we present targeted comparisons against its peers, highlighting both its strengths and persistent challenges.

Through this evaluation framework, \textsc{NoTeS-BANK} aims to serve as a diagnostic benchmark for the next generation of vision-language models in handwritten document understanding and retrieval.

\begin{table}[h]
\caption{Performance comparison of various methods. ANLS measures answer accuracy; R@1, MRR, R@5 and ACC (Global category accuracy) measure page retrieval.}
\label{tab:qa-retrieval}
\begin{adjustbox}{width=\linewidth}
\begin{tabular}{@{}lcccccc@{}}
\toprule
\textbf{Method} & \multicolumn{1}{c}{\textbf{Accuracy}} & \multicolumn{3}{c}{\textbf{Page Retrieval}} & \textbf{Domain}\\
\cmidrule(lr){2-2} \cmidrule(lr){3-5} \cmidrule(lr){6-6}
& ANLS* & R@1 & MRR & R@5 & ACC \\
\midrule
\multicolumn{6}{@{}l}{\textit{Text-based RAG}} \\
TF-IDF + LLaMa 3.1 8B &  0.0395 & 0.018 & 0.0314 & 0.058  & 5.8 \\
BM 2.5 + LLaMa 3.1 8B &  0.0721 & 0.034 & 0.0515 & 0.082 & 7.6 \\
Mp Net + LLaMa 3.1 8B &  0.0482 & 0.02 & 0.0387 & 0.076 & 6.6 \\
Minilm + LLaMa 3.1 8B & 0.0401 & 0.014 & 0.0245 & 0.038 & 8 \\

ColQwen + Qwen2VL & 0.3419 & 0.218 & 0.243 & 0.288 & 30.6\\
ColPali + Qwen2-VL 7B  & 0.3294 & 0.212 & 0.243 & 0.29 & 30.4 \\
\midrule
\textbf{Human Baseline} & 0.8667 & 0.8125 & 0.8125 & -- & 28.99\\
\bottomrule
\end{tabular}
\end{adjustbox}
\end{table}

\vspace{-0.1cm}
\subsection{Performance Trends Across Tasks}

The evaluation of models on both Evidence-Based VQA and Open-Domain QA in \textsc{NoTeS-BANK} reveals significant challenges in understanding and handling handwritten documents. While Vision-Language Models (VLMs) demonstrate promising capabilities, they still struggle with fine-grained reasoning, symbol interpretation, and multimodal retrieval. For Evidence-Based VQA, models relying solely on OCR pipelines exhibit lower IoU and ANLS scores, reinforcing the limitation of text-only processing for handwritten notes. VLMs show improved performance by incorporating visual and structural cues, but evidence localization remains a major challenge.

In Open-Domain QA, retrieval-augmented generation (RAG) models outperform traditional retrievers like BM25 and DPR, particularly in Recall@5 and NDCG@5 metrics. However, even the strongest models, such as GPT-4o RAG and Qwen-2.5-VL RAG, struggle with long-context retrieval over handwritten documents, suggesting a need for better indexing and retrieval over sparse visual information.

\vspace{0.1cm}
\noindent
\textbf{Impact of OCR on Document Understanding:} OCR-based methods perform significantly worse in both tasks, particularly for handwritten mathematical equations, symbols, and complex scientific notations. Models such as Google OCR + LLaMA 3.1 and Textract OCR + LLaMA 3.1 suffer from:
{\textit{(i)}} Loss of spatial and semantic relationships is crucial for layout-heavy content.
{\textit{(ii)}} Difficulty in transcribing non-standard handwritten characters.
{\textit{(iii)}} Inability to provide reliable evidence grounding due to segmentation errors.

\vspace{0.1cm}
This highlights the limitations of treating handwritten document QA as a text-only problem, reinforcing the necessity for joint vision-language reasoning.

\vspace{0.1cm}
\noindent
\textbf{Vision-Language Models and the Multimodal Challenge:} While closed VLMs (GPT-4o, Gemini 1.5-Pro) outperform open VLMs in answer generation, both categories struggle with localizing relevant evidence. Intern-2.5-VL and Qwen-2.5-VL show promise in handling handwritten content but fail to generalize across different domain categories.
For multimodal retrieval, OFA RAG and LLaVA RAG improve retrieval accuracy but still fail to effectively fuse retrieved document context into reasoning steps. This suggests the need for better cross-modal pretraining strategies that explicitly model symbolic and spatial dependencies.

\vspace{-0.1cm}
\subsection{Error Analysis and Limitations}

Several error patterns emerge from our analysis:

\vspace{0.1cm}

\textbf{(i)} \underline{Ambiguous Questions:} Some models fail due to question ambiguity, producing hallucinated responses instead of recognizing unanswerable questions.

\vspace{0.1cm}
\textbf{(ii)} \underline{Failure to Retrieve Key Evidence:} Even top-performing models frequently retrieve the wrong document, leading to incomplete answers.

\vspace{0.1cm}
\textbf{(iii)} \underline{Weak Layout Awareness:} Many models struggle with layout-based reasoning, especially for tables, structured lists, and diagrams.

\vspace{0.1cm}
A deeper study of failure cases in IoU-based evidence retrieval suggests that bounding-box predictions remain inconsistent, requiring improved fine-grained region detection strategies.



\section{Conclusion}

Our evaluation of state-of-the-art models on \textsc{NoTeS-BANK} highlights significant limitations in existing document QA models, particularly in handling handwritten and multimodal content. While VLMs offer promising capabilities, our results suggest that handwritten document understanding remains an unsolved challenge, requiring improvements in multimodal retrieval, evidence-based reasoning, and layout-aware processing. The findings from \textsc{NoTeS-BANK} establish a benchmark for future research, encouraging the development of more context-aware, symbolically grounded, and visually structured models. Our findings suggest several promising research directions. Future models should integrate spatially-aware tokenization to enhance symbol and diagram reasoning.
Moreover, models need better exposure to handwritten mathematical, chemical, and engineering content. Additionally, open-domain QA models should explore hierarchical retrieval approaches, combining vector-based retrieval with structural document parsing. Enhancing evidence attribution and transparency will improve trust in automatic document intelligence systems.

\bibliographystyle{ACM-Reference-Format}
\bibliography{main}

\end{document}